\pdfoutput=1

\documentclass[11pt]{article}

\usepackage[final]{acl}

\usepackage{times}
\usepackage{latexsym}

\usepackage[T1]{fontenc}

\usepackage[utf8]{inputenc}

\usepackage{microtype}

\usepackage{inconsolata}

\usepackage{xcolor}
\usepackage{graphicx}
\usepackage{subfigure}
\usepackage{tcolorbox}
\usepackage{listings}
\usepackage{float}
\usepackage{tabularx}
\usepackage{colortbl}
\usepackage{booktabs}
\usepackage{adjustbox}
\usepackage{enumitem}
\usepackage{makecell}
\usepackage{amsbsy}
\usepackage[colorinlistoftodos]{todonotes}
\usepackage{authblk}  

\definecolor{cornellred}{rgb}{0.7, 0.11, 0.11}
\definecolor{mypink}{rgb}{0.858, 0.288, 0.878}

\title{Can LLMs Learn Macroeconomic Narratives from Social Media?}

\makeatletter
\renewcommand\AB@affilsepx{, \protect\Affilfont}
\makeatother

\author[T, *]{\bf Almog Gueta}
\author[C]{\bf Amir Feder}
\author[T]{\bf Zorik Gekhman}
\author[H]{\bf Ariel Goldstein}
\author[T]{\bf Roi Reichart}

\affil[T]{Technion, IIT}\affil[C]{Columbia University} \affil[H]{Hebrew University}

\affil[*]{{\footnotesize \url{almoggu@gmail.com}}}

\begin{document}
\maketitle

\begin{abstract}\vspace{-0.9ex}

This study empirically tests the \textit{Narrative Economics} hypothesis, which posits that narratives (ideas that are spread virally and affect public beliefs) can influence economic fluctuations. 
We introduce two curated datasets containing posts from X (formerly Twitter) which capture economy-related narratives.
Employing Natural Language Processing (NLP) methods, we extract and summarize narratives from the tweets. We test their predictive power for \textit{macroeconomic} forecasting by incorporating the tweets' or the extracted narratives' representations in downstream financial prediction tasks.
Our work highlights the challenges in improving macroeconomic models with narrative data, paving the way for the research community to realistically address this important challenge. From a scientific perspective, our investigation offers valuable insights and NLP tools for narrative extraction and summarization using Large Language Models (LLMs), contributing to future research on the role of narratives in economics.\footnote{Data is available through the Social Media Archive (SOMAR): \url{https://socialmediaarchive.org/record/77}.}

\end{abstract}

\section{Introduction}\vspace{-1ex} 
Narrative Economics studies how popular narratives change over time and interact with economic behavior \citep{shiller2017narrative}. 
A key proposition within this field is that narratives can drive economic fluctuations. This is especially intriguing at the macroeconomic level, as the theory suggests that widely shared stories can influence the collective decisions of millions of individuals.
However, it presents greater challenges than for microeconomy, due to the complex interplay of numerous factors, the need for broad-covering narratives, and the inherent difficulty in isolating causal relationships.

Proving a causal link between narratives and economic changes remains challenging (\citeauthor{lucas1981rational}; an econometric identifiability problem \citep{angrist2009mostly}). Instead, we propose applying NLP methods to represent narratives extracted from social media and test their macroeconomic predictive power. 
We introduce two Twitter \citep{Twitter} datasets crafted to capture economy-related narratives (\S\ref{sec:data}), explore two NLP approaches for narrative extraction, and utilize them for prediction. First, we build prediction models directly on raw tweets, implicitly capturing the narratives within them. Second, we utilize LLMs\footnote{We refer to a variety of models as LLMs, ranging from relatively small models (e.g., BERT~\citep{devlin2018bert}) to more recent models (e.g., GPT-3.5~\citep{ChatGPT-3.5}).} to explicitly extract and summarize narratives, generating a third dataset of LLM analyses. 
We demonstrate the presence of narratives within our datasets (\S\ref{sec:textual_analysis}) and the effectiveness of our approaches in representing the aggregated economic-narrative picture.




Existing evaluation strategies for narratives' effectiveness for economic predictions are inadequate, focusing on macroeconomic correlations and anomalies, or on microeconomic predictions (\S\ref{sec:existing_efforts}). We propose utilizing downstream macroeconomic prediction tasks, where representations of latent narratives serve as input (\S\ref{sec:models_types}-\ref{sec:llms_exps}).
Interestingly, our results reveal that the successful narrative extraction we demonstrate offer only marginal improvement in macroeconomic prediction, compared to utilizing only financial information. In Section~\ref{sec:discussion} we discuss the results' implications on the validity of the Narrative Economics theory, at least when it comes to macroeconomics.
\par
Our contributions include: 
\begin{itemize}[topsep=5pt, parsep=-5pt, leftmargin=5pt]
    \item A framework for testing the narrative hypothesis on macroeconomic predictions.
    \item Three tailored datasets: two Twitter curated collections and LLM-based analyses of tweets. 
    \item Extensive analysis demonstrating the effectiveness of our methodology, revealing valuable insights while highlighting the need for new models and tasks to empirically test the Narrative Economics theory. 
\end{itemize}


\section{Preliminaries}
\subsection{Macro- and Micro- Economics}
\textbf{Macroceconomics} studies the behavior of the economy as a whole, examining factors like inflation, unemployment, and economic growth. \textbf{Microeconomics}, on the other hand, is concerned with the decision-making of individuals and firms, examining indicators like the price of a certain stock.

\subsection{Definition of Narrative}
The term "narrative" holds distinct meanings across disciplines. In Computational Narrative Understanding (CNU) \citep{chambers2008unsupervised, chambers2009unsupervised}, a narrative typically refers to a structured story with identifiable elements such as plot, characters, and setting, and the understanding of how these elements interrelate and contribute to the overall meaning and coherence of a narrative.

In contrast, our research operates within the definition of Narrative Economics. Within this framework, a narrative is a belief about the world that is shared by the population, regardless of its adherence to traditional narrative structures. This broader definition encompasses a wider range of communication phenomena, including tweets, opinions, and even rumors, that may lack the formal elements of a story in the CNU sense. These narratives can influence economic decisions and have a real affect in the world, even if they are not factually accurate. For instance, a narrative be the widespread belief that housing prices are increasing, whereas in reality (according to economic indicators) they are stagnating. This broader definition is particularly relevant to our study because we are interested in how widely held beliefs, structured in any way, can impact macroeconomic trends.
Building on this foundation, such narratives can be identified and extracted using methods such as sentiment analysis or topic modeling, which analyze the aggregate sentiment and themes present in large collections of text data to reveal prevalent beliefs.

\section{Data Resources}\label{sec:data}

\subsection{Twitter (X) Collected Datasets}\label{subsec:twitter_data} 
We use Twitter \citep{Twitter} as a source of such narratives given its real-time reflection of diverse public opinions, and collect two tweets datasets capturing economy-related narratives.
Unlike previous research focused on stock- or company- specific tweets \citep{Vamossy2021emtract, chandra2020sentiment, mengistie2021deep, sethi2020sentiment, karami2020twitter}, our approach includes a broader time-frame, diverse global topics, and comprehensive keywords to capture a wide spectrum of perspectives. 

The first dataset ranges from Twitter's early days until the COVID-19 pandemic which disrupted the economy, while the second captures more recent trends, late 2021 onward.  Both were carefully curated using targeted queries, restricted to English non-retweets with the inclusion of specific keywords, and were analyzed to ensure quality and relevance for capturing economic narratives.

\textbf{Pre-Pandemic Twitter Dataset.}\label{subsec:tweet_data_1}
We utilize Twitter API \citep{Twitter-API} to collect a comprehensive dataset of 2.4 million tweets from January 2007 to December 2020, covering six topic areas with potential economic impact (economics, business, politics, current affairs, human disasters, and natural disasters) using targeted keywords detailed in App. \ref{app:twitter_data_details}.
For each topic, we iteratively retrieved tweets for each date within the timeframe.  To prioritize viral tweets, we first retrieved the top 200 tweets based on follower count, then we randomly sampled 100 to mitigate potential bias towards highly active accounts typically associated with news outlets.
This process yielded an average of roughly 400,000 tweets per topic, contributed by about 250,000 users per topic. The average follower count of these users was 100 million. This high average reflects our initial filtering strategy, which intentionally included users with very large followings, such as those of global leaders, news outlets, and other influential figures. Minimal pre-processing was applied to the tweets, as common in literature (see \S\ref{sec:existing_efforts} and App. \ref{app:twitter_data_details}).

\textbf{Post-2021 Twitter Dataset.}\label{subsec:tweet_data_2} 
This dataset, spanning September 2021 to July 2023, was specifically constructed for LLM-driven narrative analysis (see \S\ref{subsec:textual_models}). 
To fairly test the predictive power of narratives, we needed to ensure that the employed LLM (Chat Completion API with GPT-3.5 \citep{ChatGPT-3.5}, data cutoff September 2021) relies solely on provided tweets and pre-existing world knowledge, preventing access to ``future'' information.  As our pre-pandemic collection concludes in December 2020, we assembled this new tweets dataset after the LLM's knowledge cutoff date.  

Tweets were collected monthly using Twitter Advanced Search, restricted to users with at least 1,000 followers. We focused on keywords related to business and economics topics, resulting in a diverse collection of 2,881 tweets\footnote{As this data collection method is more restricted than the previous, the resulting dataset is relatively smaller.}, 90-130 per month, tweeted by 1,255 users, including politicians, CEOs, activists, and academics. Duplicate and URL removal was applied.

\subsection{Macroeconomic Indicators}\label{subsec:financial_data}

We focus on predicting three key macroeconomic indicators:

\textbf{Federal Funds Rate (FFR):} The interest rate at which depository institutions, such as banks, lend reserve balances overnight to meet reserve requirements. The FFR serves as a Federal Reserve monetary policy tool, is influenced by public perception of economic stability, and its fluctuations impact various sectors, making it widely monitored.

\textbf{S\&P 500:} A stock market index measuring the performance of the 500 largest publicly traded companies in the U.S. It reflects collective investor confidence in economic growth and risk appetite and is widely regarded as a barometer of the overall health of the US stock market. 

\textbf{CBOE Volatility Index (VIX):} Measures market expectations of future volatility based on S\&P 500 options prices, often referred to as the ``fear gauge'' as it tends to rise during market stress and fall during market stability.

\par\vspace{4pt} These indicators are well-suited for evaluating the narratives' predictive capabilities, both in their daily frequency which aligns with Twitter tempo, and their sensitivity to public opinions and behaviours. Comparably, other widely used macroeconomic indicators have a lower frequency, such as GDP or inflation \citep{10.1257/jel.20181020, kalamara2022making, ellingsen2022news}, or are microeconomic indicators \citep{rahimikia2021realised, yang2023multi, khedr2021cryptocurrency, he2021multi}.

Detailed information on the financial indicators and pre-processing steps are in App. \ref{app:financial_data_details}.

Appendix~\ref{app:data_characteristics} summarizes all datasets and their characteristics.

\section{Experimental Setup}\label{sec:setup} 
To assess the central question, whether economic narratives can provide valuable insights for future financial movements, we design a series of prediction tasks, each aims to predict future values of one of the financial indicators as a target: FFR, S\&P 500, and VIX (described in \S~\ref{subsec:financial_data}). 
In this section we present the tasks and the evaluation methods. 

\paragraph{\large{Prediction Tasks:}} 

We assess the predictive power of narratives across three \textbf{prediction horizons}: next-day, next-week, and next-month. Given input data covering dates $T_1,...,T_t$, the model predicts values for either $T_{t+1}$, $T_{t+7}$ or $T_{t+30}$. 
Each model variant predicts a single financial indicator at a single horizon. 

    \textbf{Next value:} predicts the target's value at the specified horizon.
    
    \textbf{Percentage change:} predicts the target's percentage change between the specified horizon and the day before.
    
    \textbf{Direction change:} classifies the target's direction of change (increase or decrease) between the specified horizon and the day before.
    

These tasks are commonly used in macroeconomic research \citep{handlan2020text, 10.1257/jel.20181020, kalamara2022making, ahrens2021extracting, masciandaro2021monetary, lee2009federal, hamilton2002model, kim2023forecasting, larkin2008good}.

\paragraph{\large{Evaluation:}}
We evaluate our models using Mean Squared Error (MSE) for the regression tasks, and Accuracy and $F_1$-Score for classification. Additionally, we compare our models against two types of baselines described below.
In all experiments besides \S~\ref{subsec:llm_exp}, where the model is not being trained, we utilize an 80-20 train-test split, adhering to chronological order to preserve temporal context.

\paragraph{\large{Financial Baselines:}}\hfill











\begin{table}[h!]
\small
\setlength{\tabcolsep}{5pt}  
\begin{tabularx}{\columnwidth}{l|l|c} 
\toprule
\textbf{Baseline} & \textbf{Description} & \textbf{Type} \\ 
\midrule
As-previous & Next value is the same as prev. & C/R \\
Inverse-previous & Next value is inverse of prev. & C \\
Week-majority & Majority vote of previous week & C \\
Train-majority & Majority vote of training data & C \\
Up-predictor & Always predicts an "increase" & C \\
Down-predictor & Always predicts a "decrease" & C \\
Train-mean & Mean value of training data & R \\
\bottomrule
\end{tabularx}
\caption{Financial baselines. C: Classification, R: Regression}
\label{tab:financial_baselines}
\end{table}


\paragraph{\large{Counterfactual Textual Baselines:}}\label{subsec:textual_baselines}\hfill

\noindent\textbf{Random texts:} To evaluate whether the LLM actually utilizes the content of tweets, we establish a baseline which feeds it with randomly generated sentences comprised of varying random words.

\noindent\textbf{Shuffled tweets:} We assess model reliance on temporal narratives by feeding the LLM with chronologically disordered tweets, to isolate the impact of temporal narratives from confounding patterns or memorization. A well-functioning model should outperform this baseline, indicating its reliance on relevant temporal narratives for prediction.

\noindent\textbf{Synthetic ``narratives'':} We generate synthetic narrative-like sentences expressing positive or negative cues, aligned with subsequent changes in the financial indicator. This allows us to assess the model's ability to infer relationships between aligned narratives and the following market changes.

\section{Models}\label{sec:models_types}
Our models are categorized into 3 categories, based on the signals they use as input:

    \textbf{Financial (F):} utilizes historical financial data, from the past week or month. 
    
    \textbf{Textual (T):} leverages solely textual data, either raw tweets or tweets' analyses. 
    
    \textbf{Textual \& Financial (TF):} draws upon both textual and financial data as input. We aim to effectively utilize both insights from textual narratives and historical financial patterns for enhanced prediction accuracy.
Outperforming an \textit{F} model with a \textit{TF} model can demonstrate the added value of textual narratives in enhancing prediction capabilities.

\vspace{4pt} 
Our model selection progresses from simpler models, commonly used in the financial literature (see \S\ref{sec:existing_efforts}), to more advanced architectures. This progression serves two purposes: (1) Achieving positive results with simpler models provides a stronger indication for the predictive signal of narratives; and (2) It allows us to build upon existing research in Narrative Economics, primarily rooted in finance and often utilizes relatively simple models, before exploring more advanced NLP approaches.

\subsection{Financial Models}\label{subsec:financial_models}
The financial models include traditional \textit{machine learning (ML) models} (Linear Regression, Lasso, Ridge, SVM, Random Forest, and Logistic Regression), \textit{DA-RNN} \citep{qin2017dual}, and \textit{T5} \citep{raffel2020exploring}. 
Each model processes a sequence of historical financial values, either as individual features or as a time-series. T5 receives these inputs in textual form (``increase'' and ``decrease'') for ``direction change'' classification.

\subsection{Textual Models} \label{subsec:textual_models}
\begin{figure}[h!]
\centering
\includegraphics[width=1\columnwidth]{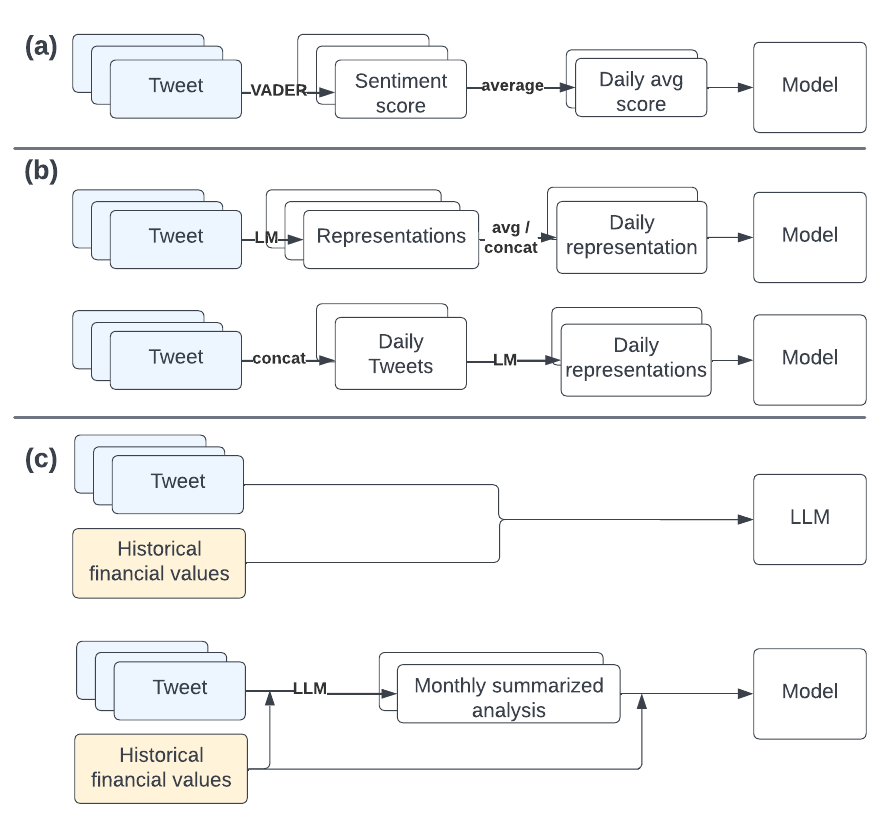}
\caption{Textual models' pipelines to represent textual data as part of the prediction pipeline. \textbf{(a)} Daily sentiments. \textbf{(b)} Individual and joint tweets LLM's representations. \textbf{(c)} LLM analysis for prediction and as input to a subsequent prediction model.}	
\label{fig:textual_pipelines}	
\end{figure}

We employ increasingly complex text representations, serving as features for the remaining of the prediction model, as illustrated in Figure~\ref{fig:textual_pipelines}.

    \textbf{Daily Sentiments:} Motivated by literature suggesting that aggregated sentiments capture the core of textual narratives and enhance economic prediction \citep{macaulay2023narrative, yang2023multi, adams2023more, kim2023forecasting, gurgul2023forecasting}, we represent each tweet with its sentiment score utilizing VADER \citep{hutto2014vader}, given its extensive adoption in the financial literature \citep{kalamara2022making, khedr2021cryptocurrency, kim2023forecasting}, providing a basis for comparison with existing research. Daily sentiment scores are then computed by averaging individual tweets scores within each day, and concatenated over a week. 
    
    \textbf{Embedding-Based Representations:} 
    We utilize pre-trained language models (BERT \citep{devlin2018bert}, RoBERTa \citep{liu2019roberta}, and T5 \citep{raffel2020exploring}) to derive embeddings for individual or concatenated tweets. 
    \textit{Individual tweet} embeddings (CLS token or averaged word embeddings) are aggregated daily by averaging or concatenating embeddings of same-date tweets. \textit{Joint tweets} embeddings encode together multiple concatenated tweets from the same date, potentially capturing their collective meaning without explicit aggregation, avoiding potential information loss.

    \textbf{LLM-Generated Analyses (hybrid model):} We create concise monthly analyses by feeding a month of tweets from the Post-2021 Twitter dataset (\S\ref{subsec:tweet_data_2}) and corresponding financial values of the target indicator to OpenAI's Chat Completion API, GPT-3.5 \citep{ChatGPT-3.5} (see App. \ref{app:llm_method} for LLM details and a prompt example).
    These analyses are either used directly for prediction (\S\ref{subsec:llm_exp}) or as an input to a subsequent T5 model (\S\ref{subsec:llm_t5_exp}).  
    Notably, the LLM receives both textual and financial inputs to enable analyzing relationships. Thus, this method applies only for the \textit{TF} model type.
    

\subsection{Integrating Textual and Financial Models}\label{subsec:t_f_integration}
Having established dedicated models for processing \textit{T} and \textit{F} data, we now address the strategies for combining these representations in \textit{TF} models to produce unified predictions:

    \textbf{Concatenation:} The simplest approach is concatenating the \textit{T} and \textit{F} representations. 
    
    \textbf{DA-RNN \citep{qin2017dual}:} The dual-stage attention-based RNN model, which was used in related work \citep{wu2018hybrid, 8679218, 9718424}, predicts the current value of a time-series based on its previous values and those of exogenous series. We feed historical financial representations (\textit{F}) as the time series and textual representations (\textit{T}) as the exogenous series. 
    
    \textbf{Prompt-based integration:} Both LLM-generated analyses (\textit{T}+\textit{F}) and raw financial values (\textit{F}) are provided to a T5 model \citep{raffel2020exploring} as separate segments with leading prompts instructing the model on how to use each data source (see App. \ref{app:t5_prompt} for a prompt example). 
\par Given a \textit{TF} model, we can derive \textit{T} or \textit{F} models by selectively omitting or zeroing either the \textit{F} or \textit{T} features, respectively.


\section{Analyzing Narratives in Textual Data}\label{sec:textual_analysis}
Since our datasets serve as the input sources for the downstream prediction models, we first assure the presence of latent narratives within them.

\subsection{Tweets Narrative Analysis} 

\begin{figure}
\centering
\includegraphics[width=1\columnwidth]{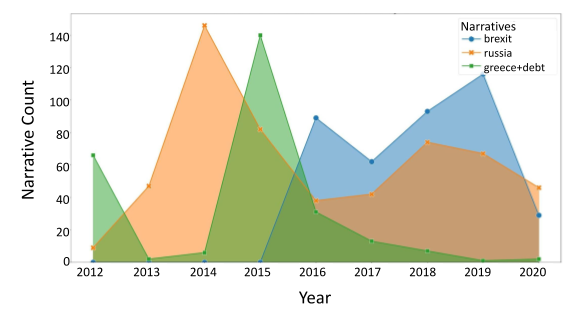}
\vspace{-25pt}
\caption{Temporal distribution of top three narratives from the ``economics'' dataset, extracted by RELATIO \citep{ash2021relatio} (see \ref{sec:existing_efforts}): UK's Brexit, Greece's financial debt, and Russia's financial crises. We can see the evolving nature of these narratives over time, where the distribution is aligned with real-life related events.}	
\label{fig:relatio_narratives}	
\end{figure}

We analyze the narratives captured in our two Twitter corpora (described in \S\ref{subsec:twitter_data}) using RELATIO \citep{ash2021relatio}, an algorithm that extracts political and economic narrative-tokens from text (see \S\ref{sec:existing_efforts}). 
Figure \ref{fig:relatio_narratives} shows the temporal distribution of the three most frequent narratives in the economy-topic dataset: UK's Brexit, Greece's financial debt, and Russia's financial crises. It demonstrates the evolving nature of these narratives over time, with peaks aligning with related real-world events such as the debt crises Greece experienced in 2015 and the UK's referendum in June 2016 about leaving the European Union. 
The presence of these narratives and their temporal dynamics within our dataset confirm the potential of social media data for economic understanding and forecasting. Moreover, the rapid shifts in word frequency following significant events suggest that economically relevant events are swiftly reflected in our data, making it a valuable source for short-term economic prediction.
For further analyses, see App. \ref{subapp:twitter_analysis}.

\subsection{LLM-Based Narrative Analysis}\label{subsec:llm_analysis}
Having demonstrated the presence of narratives within our raw data, we now analyze the LLM's ability to extract and summarize them.
Following the methodology in \S~\ref{subsec:textual_models}, we generated 697 LLM-based analyses, one for each date from September 2021 to July 2023. 
The outputs generated by the LLM can be divided into two main components: the analysis of tweets, and the potential impact of them on the financial target. In each component, the analyses contain \url{~}9 sentences on average. 

A snippet of LLM-generated analysis is presented in Figure~\ref{fig:snippet_llm_analysis_example_1} (see full examples in App.~\ref{subapp:llm_analysis}).
This snippet illustrates the LLM's information aggregation, summarizing and distinguishing between opinions and events expressed in the tweets. Furthermore, the LLM connects the insights it found to potential future impacts on the financial indicator, a crucial first step towards prediction.

\begin{figure}[h]
    \centering
    \begin{tcolorbox}[width=1\columnwidth]
\textbf{Analysis of Tweets:}

...Some tweets express concerns about inflation, rising interest rates, and the impact on the economy and personal finances...Several tweets highlight the impact of government policies on various sectors, such as healthcare, student loans, and housing.
Some tweets express skepticism towards central banks and their role in the economy...A few tweets discuss the impact of global events, such as the Russian mobilization...

\textbf{Potential Effect on S\&P 500:}

...Concerns about inflation, rising interest rates, and economic instability expressed in the tweets may lead to increased market volatility and potential declines in the S\&P 500...Global events mentioned in the tweets...may have indirect effects on the S\&P 500 through their impact on global markets and investor sentiment.
    \end{tcolorbox}
    \vspace{-15pt}
    \caption{Snippet of LLM-based analysis for 29/08/2022 to 28/09/2022. In this time period the Federal Reserve raised the interest rates in an effort to combat inflation, the US Supreme Court ruled that the Biden administration could not extend the pause on student loan payments, and more. See full analysis in App.~\ref{subapp:llm_analysis}}
    \label{fig:snippet_llm_analysis_example_1}
\end{figure}

\section{Predicting using LLM representations}\label{sec:lm_exps}
After analyzing our textual data and demonstrating the existence of narratives along with their temporal shifts aligned with world events, this section delves into employing LLMs to represent these narratives in order to use them for financial forecasting.


\subsection{Sentiment-Based Next-Day Prediction}\label{subsec:sentiment_exp} 
\textbf{Method.} 
We use daily sentiment from the past week as features (see \S~\ref{subsec:textual_models}) for FFR next-day prediction. Sentiments are fed as features into several classical ML models, described in \S~\ref{subsec:financial_models}, applied to ``next value'' and ``direction change'' tasks. 
For the ``prediction change'' task the financial input is encoded either as numerical features or binary increase/decrease values to align with the label. 


\textbf{Results.} 
Table~\ref{tab:vader_ml} presents the results of the most performant models (full results are in App.~\ref{app:sent_exp}). 
In classifying ``direction change'', models with financial input (\textit{F} \& \textit{TF}) outperform text-only models (\textit{T}), exhibiting a 5\% accuracy improvement (0.94 vs. 0.89).
The slight difference between \textit{F} and \textit{TF} models (0.936, 0.939) suggests text has little impact.\footnote{Feature importance confirms this, with the previous day's financial feature dominating in \textit{TF} models (not shown).} The \textit{T} models achieve comparable accuracy to the \textit{F} baselines (0.89, 0.81).
In predicting ``next value'', best \textit{F}, \textit{TF} and \textit{T} models yield comparable MSE to the ``train-mean'' baseline (15.4, 15.6).\footnote{Feature importance reveals low scores for both sentiments and financial features (not shown).} 

\textbf{Takeaway.} Sentiments lack nuances needed for financial prediction, and classical ML models have limited capabilities. Several models failed to surpass the performance of the non-learned baselines, indicating the necessity for improved text representation and more advanced prediction models.

\begin{table}[h]
\small  
\setlength{\tabcolsep}{8pt}  
\begin{tabularx}{\columnwidth}{l|l|c} 
\toprule
\textbf{Type} & \textbf{Model}& \textbf{Accuracy}\\
\midrule
F baselines & As-previous & 0.812\\
F & Random Forest Numeric & \bfseries{0.936}\\
TF & Random Forest Numeric & \bfseries{0.939}\\
T & Logistic Regression & 0.885\\
T & SVM & 0.885\\
\midrule
\midrule
\textbf{Type} & \textbf{Model}& \textbf{MSE}\\
\midrule
F baseline & Train-mean & 15.661 \\
F & SVM & 15.416 \\
TF & SVM & 15.416 \\
T & SVM & 15.36 \\
\bottomrule
\end{tabularx}
\caption{Results of most performant models for predicting FFR using sentiments. \textbf{1.} In ``direction change'', financial features are encoded either numerically. \textit{TF} and \textit{F} perform nearly identically, outperforming \textit{T} models and \textit{F} baselines. \textbf{2.} In ``next value'', best models are comparable to each other and to the \textit{F} baseline.}
\label{tab:vader_ml}
\end{table}

\subsection{Embeddings for Time-Series Prediction}\label{subsec:emb_exp}
\textbf{Method.} To better capture the richness and complexity of information concealed within the tweets, we turn to the two ``Embedding-Based Representations'' described in \S~\ref{subsec:textual_models}. 
Additionally, we step beyond ML models and instead utilize the time-series deep learning model DA-RNN \citep{qin2017dual}, designed to capture temporal dynamics and complex relationships within data (see \S~\ref{subsec:t_f_integration}).


Through rigorous evaluation, we explore various model configurations, target variables (FFR and VIX), tasks (``percentage change'', ``direction change'' and both together), prediction horizons, LLM architectures, aggregation methods, and the daily number of tweets.
Lastly, we assess the models' reliance on temporal context and relevant narratives, using the ``random texts'' and ``shuffled tweets'' counterfactual baselines described in \S~\ref{subsec:textual_baselines}. These baselines are integrated with the F models, as explained in \S~\ref{subsec:t_f_integration}, to create TF baselines. \newline
Due to space constraints, we present results for a single task and setting- VIX ``next value'' prediction with 1- or 7- days horizons. Additional experiments showed a recurring pattern to the presented results. 

\begin{figure}[h]
\centering
\includegraphics[width=1\columnwidth]{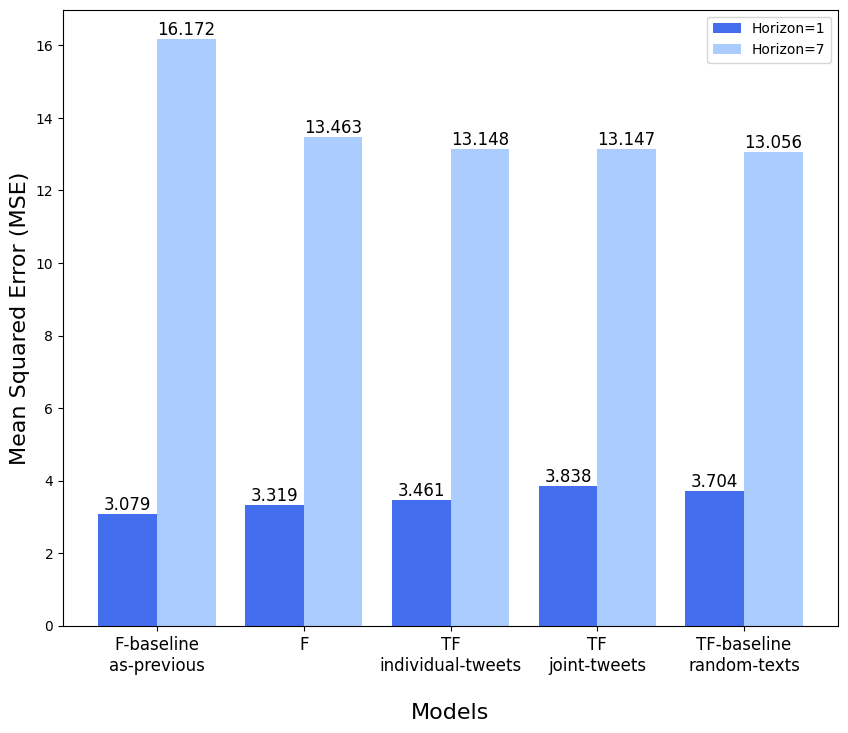}
\vspace{-20pt}
\caption{VIX ``next value'' prediction for 1/7-days horizons. The \textit{F} and \textit{TF} baselines outperform all models in 1- and 7-day horizons, respectively, suggesting all models struggle to learn from tweets for the prediction.}	
\label{fig:lm_results}	
\end{figure}

\textbf{Results.} Figure~\ref{fig:lm_results} presents the models' performance. For a 1-day horizon, the ``as-previous'' \textit{F} baseline outperforms all other models (3.079 MSE). This suggests that the input information might not be beneficial for such a short-term prediction. 
For a 7-day horizon, both \textit{TF} models (13.148, 13.147) initially appear to outperform the \textit{F} model (13.463) and \textit{F} baseline (16.172), implying a potential influence of the textual content. However, the ``random texts'' \textit{TF} baseline outperforms (13.056) all other models, indicating that the good performance of the \textit{TF} models is not likely due to the content of the tweets. We hypothesize that the presence of text improves performance, even when random, due to either spurious correlations or random noise that helps the model generalize, similarly to regularization techniques. 
The difficulty in capturing and representing the aggregated tweets information in a way that is meaningful for financial prediction might be a contributing factor. Additionally, challenges may lie in using historical financial data to predict future values of an indicator characterized by frequent random movements and fluctuations.


\textbf{Takeaway.} The models struggle to learn from the tweets for the macroeconomic predictions, suggesting that implicitly representing and aggregating latent narratives within LLMs remains challenging.

\section{Predicting using LLM Analyses}\label{sec:llms_exps}
Our former attempts to predict financial indicators directly from raw tweets proved insufficient, potentially due to the difficulty of implicitly utilizing narratives and aggregating information from diverse tweets. To address this, we apply generative LLMs, providing them tweets and financial data from the same time period, to summarize insights and analyze potential effects on future market movements.
We initially explored using the LLM directly as a predictor (\S\ref{subsec:llm_exp}), but due to limited success, we repurposed its analyses as refined inputs for a subsequent prediction model (\S\ref{subsec:llm_t5_exp}).

\subsection{LLMs for End-to-End prediction}\label{subsec:llm_exp} 


We predict the average weekly VIX (or S\&P 500) value based on a monthly window of tweets and corresponding VIX  (or S\&P 500) values.
Utilizing the web chat version of GPT \citep{gpt-chat} for continuous reasoning, we tested few- and zero-shot settings, and multi-step reasoning with Chain-of-Thought (CoT) prompting \citep{wei2022chain}.
See App.~\ref{app:llm_as_predictor} for full description.

Overall, the LLM consistently produced meaningful narrative analyses and comprehension of financial implications, although not being instructed to do so, but exhibited inconsistencies in integrating these insights for prediction. For example, it occasionally refused to provide predictions due to safety guardrails. In other cases it mirrored input ranges neglecting the potential impact of successfully analyzed narratives.
When presented with ``synthetic narratives'' (\S~\ref{subsec:textual_baselines}), it recognized the change direction but struggled to quantify the magnitude of it.
Addressing these limitations could unlock the full potential of LLMs for financial forecasting.

\subsection{Two-Stage prediction with LLM Analyses}\label{subsec:llm_t5_exp}
\textbf{Method.} The previous experiment revealed the LLM's ability to create insightful analyses of tweets-financial data (see \S\ref{subsec:llm_analysis} and \ref{subapp:llm_analysis} for LLM-outputs analysis). Here we use these analyses as inputs for a dedicated prediction model (see \S\ref{subsec:textual_models}), to predict the S\&P 500 ``direction change''.
Unlike the embedding-based approach (\S~\ref{subsec:emb_exp}) which struggled to aggregate diverse narratives, here the LLM produces concise analyses. Compared to the LLM predictor approach (\S~\ref{subsec:llm_exp}), here we implement a separate fine-tuned model for the downstream prediction task. Additionally, we compare the models to the ``synthetic narratives'' \textit{T} baseline (see \S~\ref{subsec:textual_baselines}). 



\textbf{Results.} Table~\ref{tab:t5_with_llm} shows that there is no significant difference between the best \textit{TF} and \textit{F} models, with a performance gap of \url{~}2\% on a limited test set of \url{~}90 samples.\footnote{As a reminder, we can only use the second Twitter dataset, of tweets that were posted after the LLM's training cutoff date, and our financial indicators are of daily frequency, therefore the small dataset for this type of experiments.}
The McNemar's test \citep{P18-1128} reveals no statistically significant difference (p-value=0.48).
We regard the small gap as a negative result and understand if this causes curiosity of the readers.
Notably, in comparison to previous experiments, here the models surpass all baselines.





\textbf{Takeaway.} While \textit{TF} and \textit{F} models outperform all others, the gap between them is insignificant.

\begin{table}[htbp]
\small  
\setlength{\tabcolsep}{5pt}  
\begin{tabularx}{\columnwidth}{l|l|c|c} 
\toprule
\textbf{Type} & \textbf{Model} & \textbf{Accuracy} & $\mathbf{F_1}$\textbf{-Score} \\
\midrule
 & Train-majority & 0.424 & 0.0  \\
 & Week-majority & 0.484 & 0.598 \\
F- & As-previous & 0.484 & 0.552 \\
baselines & Inverse-previous & 0.517 & 0.511 \\
 & Up-predictor & 0.576 & 0.731 \\
 & Down-predictor & 0.424 & 0.0 \\
\hline
F & T5 Base & \bfseries{0.604} & 0.723  \\
F & T5 Large & 0.593 & \bfseries{0.727} \\
\hline
TF & T5 Base & 0.626 & 0.738 \\
TF & T5 Large & \bfseries{0.627} & \bfseries{0.742} \\
\hline
T & T5 Large & 0.587 & 0.726 \\
\hline
T-baseline & Synthetic narratives & 0.489 & 0.254 \\
\bottomrule
\end{tabularx}
\caption{S\&P 500 ``direction change''. \textit{TF} and \textit{F} models outperform all other models, with an insignificant gap of \url{~}2\% favoring the \textit{TF} model.}
\label{tab:t5_with_llm}
\end{table}

\section{Related Work}\label{sec:existing_efforts}
\textbf{Financial Market Prediction.} 
Financial prediction is a longstanding research area, with approaches ranging from classical quantitative methods \cite{arthur1995complexity, andersen1999forecasting, 10.1257/0895330041371321, hamilton2002model, athey2019machine} to traditional ML models using historical financial data \cite{kalamara2022making, 10.1257/jel.20181020, masciandaro2021monetary}. However, these methods often struggle to capture market complexities due to their limitations in modeling non-linear relationships.

Others leverage RNNs, CNNs, and attention mechanisms  \cite{yang2023multi, handlan2020text, lee2009federal, he2021multi, qin2017dual}. Very large LLMs have recently started blooming in this domain, for learning from financial texts and time-series data  \cite{garza2023timegpt, wu2023bloomberggpt, yang2023fingpt, yu2023temporal, xie2023pixiu} and for stock price prediction \cite{swamy2023llm, chen2023chatgpt, lopez2023can, koa2024learning}. Yet, their often closed-source nature and tendency to hallucinate limit their application \cite{bybee2023surveying}.

Features used for prediction include a wide range of financial information, from stocks and macroeconomic indicators to news, Fed announcements and companies reports \citep{ahrens2023mind, larkin2008good, yang2023multi, rahimikia2021realised, kim2023forecasting, 10.1257/jel.20181020}.

\textbf{Social Media and Finance.}
Social media provides a wealth of user-generated content for analyzing investors opinions and market dynamics. Researchers commonly collect data by filtering hashtags or accounts, often preprocessing it by removing emojis and links \citep{Vamossy2021emtract, chandra2020sentiment, mengistie2021deep, sethi2020sentiment, karami2020twitter}. 

Studies have explored various text representations, from closed forms such as sentiments, emotions and topics \citep{wang2023deepemotionnet, mengistie2021deep, wazery2018twitter, khan2020big, kaur2020sentiment, sethi2020sentiment, soumya2020sentiment}, to embeddings from pre-trained LLMs \citep{sethi2020sentiment, soumya2020sentiment, kalamara2022making, 10.1257/jel.20181020, chanda2021efficacy, ye2020document} or specific-domain fine-tuned LLMs \citep{kim2023forecasting, chu2023language, gurgul2023forecasting}. 

Common approaches to aggregating these representations, such as averaging, extracting topics, or creating indices \citep{kalamara2022making, ellingsen2022news, mezo2021text, adams2023more, kim2023forecasting, khedr2021cryptocurrency} often struggle to capture the diversity of opinions, topics and narratives. We propose generating more holistic LLM analyses, aggregating common narratives while distinguishing differing viewpoints.

Social media data has been used to learn relationships with financial markets, often finding correlations and similarities  \citep{chandra2020sentiment, karami2020twitter, nyman2021news, gholampour2019exchange, ahrens2021extracting, masciandaro2021monetary, macaulay2023narrative}.
While this data shows promise for financial prediction, existing work often relies on simplified representations, focuses on microeconomic variables or events-predicting \citep{gurgul2023forecasting, adams2023more, wang2023deepemotionnet}, while macroeconomic forecasting is underexplored.



\par\textbf{Learning Narratives.}
Emerging research explores Narrative Economics \citep{ellingsen2022news, nyman2021news, ahrens2021extracting, macaulay2023narrative, feder2022causal}. For example, RELATIO \citep{ash2021relatio} is an algorithm extracting political and economic narratives from texts by identifying entity groups and mapping their relationships. While useful for analysis, RELATIO outputs discrete tokens, making them less suitable for our LLM-based prediction tasks.


\par \textbf{In summary}, while significant progress has been made in leveraging social media data and NLP models for economic applications, the impact of narratives on macroeconomic movements remains unclear. This work builds upon existing research, but distinguishes itself by combining advanced NLP, for extracting and aggregating a broad scope of economy-related narratives, and integrating them with financial data to assess their potential to improve short-term macroeconomic predictions, an intersection not yet addressed in the literature.

\section{Discussion and implications of Results}\label{sec:discussion}


This research explores the Narrative Economics hypothesis, extracting and analyzing economy-related narratives from social media and testing their utility for downstream macroeconomic prediction, with tweet representations serving as input.

While demonstrating the presence of narratives within our curated datasets and establishing NLP building blocks for narrative extraction, evaluating their macroeconomic impact remains a challenge. 
Our models incorporating narrative data showed limited improvement over those using solely financial data. They failed to consistently outperform our baselines or financial models, and any observed improvements were marginal and statistically insignificant and we regard it as a negative result.

Addressing the question possessed in the title— \textbf{Can LLMs Learn Macroeconomic Narratives from Social Media?}—our findings suggest so, as demonstrated by our analyses (\S\ref{sec:textual_analysis}). Yet, the practical utility of this learned knowledge for economic applications remains an open question.
This gap between successful extraction of narratives and limited improvement in macroeconomic prediction raises a question regarding the extent to which these narratives, on their own, can truly drive and predict economic fluctuations, at least at the macroeconomic level.
This study serves as a foundation for further exploration, highlighting the need for new macroeconomic models or tasks designed to assess the extracted narratives' influence on the economy.

\section{Limitations}\label{sec:limitations}
This research has potential limitations while facing several challenges. 
First, focusing on short-term prediction horizons (nowcasting) presents a significant challenge due to the inherent complexity and randomness of economic markets. The Efficient Market Hypothesis suggests that the predictive power of nowcasting is limited, as public information is instantly reflected in asset prices. However, Narrative Economics proposes that narratives can affect peoples' decisions and therefore help us predict and understand economic fluctuations \citep{shiller2017narrative}.

We focused on a limited number of economic targets (FFR, S\&P 500, and VIX) influenced and shifted by diverse, external and unobserved sources. Utilizing other financial targets or other tasks, such as anomaly detection or profit prediction, might have lead to stronger evidences of the impact of narratives on the economy.

Identifying the temporal lag between the emergence of a narrative and its potential influence on financial targets presents another hurdle. Although being comprehensive, our experiments only examined limited lags and prediction horizons. 

While our datasets were carefully curated to capture potential narratives, identifying them definitively is challenging, especially when aggregating multiple narratives for a holistic economic picture. The definition of ``narrative'' is broad and subjective, and narratives are typically only recognizable as such in retrospect. Combined with the inherent noise and susceptible to misinformation in social media, extracting reliable narratives with certainty is a complex task.

Regarding geographical aspects, this work is limited to English-language data, and to US-centric macroeconomic indicators.

Lastly, we are limited to publicly accessible LLMs with a known cutoff date, to avoid potential ``future'' world knowledge. Utilizing other models might lead to better results.

It is important to acknowledge that this research deals with predicting human and economic behavior, which carries potential risks of misuse. This technology could be applied in harmful or unfair ways, and therefore should be developed and used with caution and awareness of its ethical implications.





\bibliography{acl_latex}

\appendix
\onecolumn

\section{Twitter Datasets Details}\label{app:twitter_data_details}

\textbf{Twitter topics' keywords: } Table \ref{tab:twitter_topics_keywords} represents the keywords used to collect the Twitter datasets. Each topic has a corresponding list of keywords related to the topic. See Section~\ref{sec:data} for further explanation.

\begin{table}[!htb]
\centering
\begin{tabularx}{\linewidth}{|l|X|}
\rowcolor{lightgray} \multicolumn{1}{c}{Topic} & \multicolumn{1}{c}{Key Words} 
\tabularnewline 
Politics & politics, enough, occupy, coup, demonstration, protest, protesters, corruption, active, embassy, government                                                   
\tabularnewline \midrule
Business & retail, business, business owner, business relationship, ecommerce, entrepreneurs, entrepreneurship, CEO, management, invest, founder, innovation, patent, economic, finance, economy, financial, funding, stock
\tabularnewline \midrule
Economy & macroeconomics, yield, inflation, mortgage, recession, debt, interest rate, loan 
\tabularnewline \midrule
Current Affairs & human interest, news, all news, breaking news, whats happening, notify, news desk, world news                                                     
\tabularnewline \midrule
Natural Disasters & global disaster, natural disaster, world, evacuation, hit, shaking, aftershock, safe, disaster, evacuate, earthquake, tsunami, flooding, volcano, eruption, massive, damage, destroyed                                                           \tabularnewline \midrule
Human Disasters & explosion, terror, attack, horror, crash, shot, shots, shooting, terrible, shocking, police, killed, national military, national security, national terrorism
\tabularnewline \bottomrule
\end{tabularx}
\caption{Collected topics and their corresponding keywords.}
\label{tab:twitter_topics_keywords}   
\end{table}

\textbf{Additional Keywords for the Post-2021 Twitter Dataset:} occupy, invest, business, economic, finance, economy, financial, stock, macroeconomics, yield, inflation, mortgage, recession, interest rate, loan, GDP, unemployment, federal reserve, fed, exchange rate, monetary policy, FOMC, central bank.

\paragraph{Data Pre-Processing: } The tweets underwent minimal pre-processing including emoji-to-text conversion, duplicates removal, links removal, spelling corrections, lemmatization and tokenization.

\section{Financial Indicators Details}\label{app:financial_data_details}

\begin{figure*}[!h]
\centering
\includegraphics[width=1\textwidth]{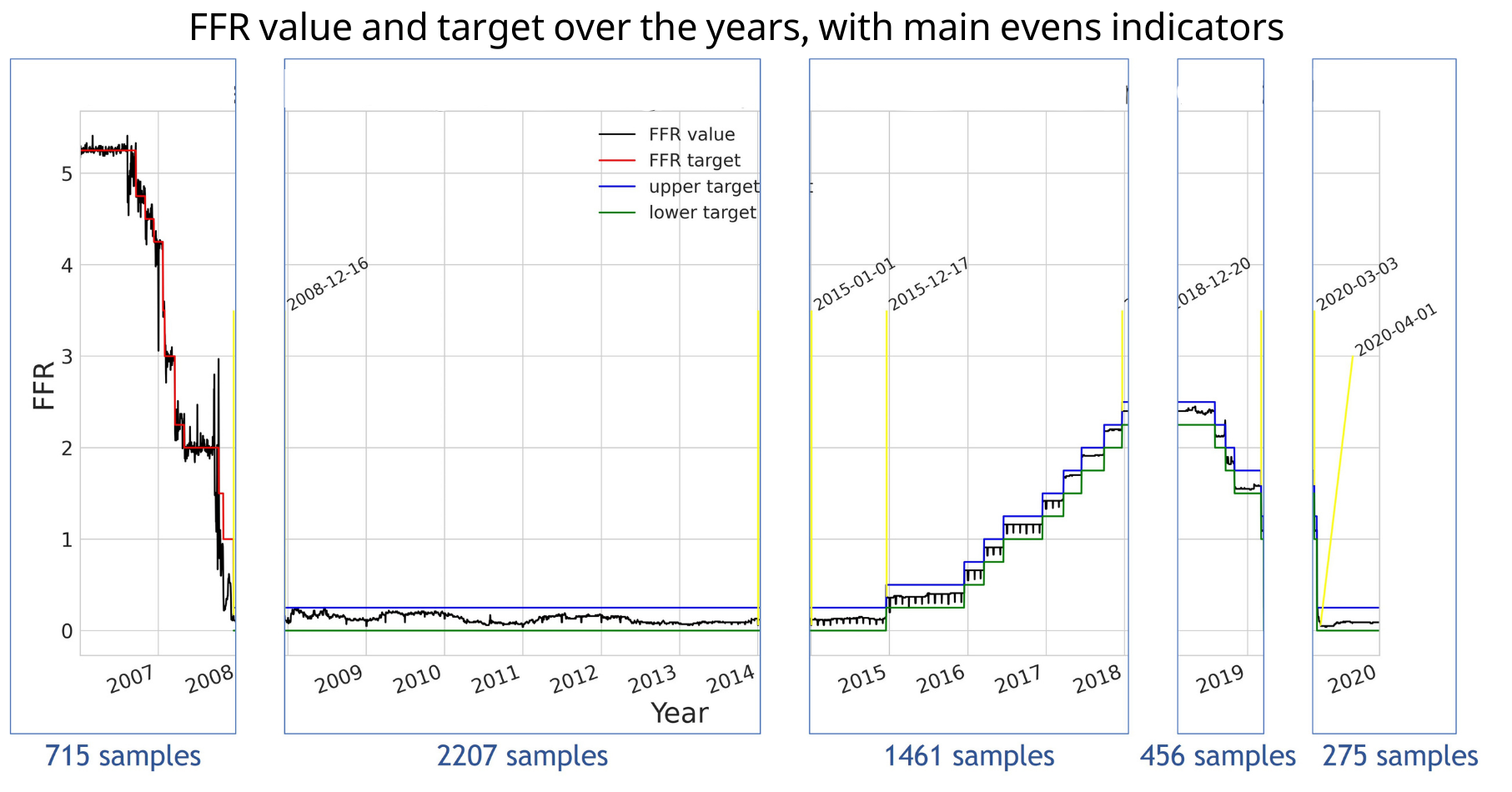}
\caption{Blocking split of FFR data, isolating time-periods with consistent distributions.}	
\label{fig:ffr_data_splits}	
\end{figure*}

\paragraph{Federal Funds Rate (FFR):} The FFR is the interest rate at which depository institutions, such as banks, lend reserve balances to each other overnight  to meet reserve requirements.
The FFR serves as a critical monetary policy tool employed by the Federal Reserve to influence economic activity by adjusting the target range of the FFR in quarter-point increments following Federal Open Market Committee (FOMC) meetings.
Changes in the FFR can have significant impacts on various sectors of the economy, including consumer spending, investments, housing, and financial markets.
Changes in the FFR can have significant impacts on various sectors of the economy, including consumer spending, investments, housing, and financial markets, making it a widely monitored indicator of overall economic health.
The dataset was downloaded from the publicly available economic data at FRED® website \cite{fred}.
To address historical distribution changes, mainly due to the Great Recession and the COVID-19 pandemic, we employ ``blocking split for time-series data'', separating periods with consistent distributions. The longest block (December 2008-January 2015) yields 2207 samples and was used in experiments. Figure~\ref{fig:ffr_data_splits} shows the blocking splits of the FFR. The second block (December 2008-January 2015) is the longest and was used for experiments.

\paragraph{S\&P 500:} This stock market index measures the performance of the 500 largest publicly traded companies in the US. It is widely regarded as one of the best indicators of the overall health and performance of the US stock market and, by extension, the broader economy. Changes in the index value can indicate shifts in investors' narratives, confidence in economic growth prospects, risk appetite, and expectations for future economic conditions. This dataset was downloaded from Yahoo Finance® website \citep{yahoo}. We utilize the closing price of the asset. Additionally, to mitigate the impact of the COVID-19 pandemic, we only utilize post-pandemic data.

\paragraph{CBOE (Chicago Board Options Exchange) Volatility Index (VIX):} The VIX measures market expectations of future volatility based on S\&P 500 options prices. It reflects investors' perceptions of the risk and uncertainty in the stock market over the next 30 days. The VIX is known as the ``fear gauge'' as it tends to rise during times of market stress and uncertainty and fall during periods of market stability. Analyzing changes in the VIX provides insights into investors' sentiments, risk appetite, and economic confidence. This dataset was downloaded from Yahoo Finance® website \citep{yahoo}. Similar to the S\&P 500, we utilize the close price of the index, either pre- or post-pandemic data, avoiding their integration within the same model.

\section{Datasets Characteristics}\label{app:data_characteristics}
Table~\ref{tab:data_characteristics} summarizes all datasets and their characteristics. Each dataset has a different role in the experiments. The full descriptions of the datasets are presented in Section~\ref{sec:data}.

\begin{table}[!h]
\centering
\begin{tabularx}{\linewidth}{|c|X|X|p{8cm}|}
\toprule
\rowcolor{lightgray} Dataset Type & Dataset Name & Time Span & Characteristics 
\tabularnewline\midrule
Twitter & Pre-Pandemic Twitter Dataset & 2007-2020 & Collected daily using 6 topics, minimal followers threshold  
\tabularnewline\midrule
Twitter & Post-2021 Twitter Dataset & 2021-2023 & Collected monthly using a single topic, minimal followers threshold, tweets poster after training cutoff of the used LLM 
\tabularnewline\midrule
Financial & Federal Funds Rate & 2007-2020 \newline (2008-2015) & Daily macroeconomic indicator, impacts various sectors, widely monitored 
\tabularnewline\midrule
Financial & S\&P 500 & 2021-2023 & Daily macroeconomic indicator, its fluctuations indicate shifts in investors' narratives 
\tabularnewline\midrule
Financial & Volatility Index & 2007-2020 \newline 2021-2023 & Daily macroeconomic indicator, reflects investors' uncertainty of the market 
\tabularnewline\bottomrule
\end{tabularx}
\caption{Summary of datasets with their characteristics.}
\label{tab:data_characteristics}   
\end{table}

\section{LLM Details}\label{app:llm_method}
Here we describe implementation details for our LLM-generated analyses pipeline. This Appendix extend the information in Section~\ref{subsec:textual_models}. We elaborate on the prompt and parameters, and present a prompt example.

\textbf{Prompt:} 
The prompt of the LLM contains instructions, a set of tweets from the analyzed month, and a set of the financial indicator values (VIX or S\&P 500) from the same time period. We feed the LLM with both tweets and indicator values to allow it to potentially understand connections or correlations between them while generating the output.
We limit to up to 10 tweets per day to avoid a long prompt. 
The sets of tweets and financial values are given in a dictionary format keyed by their date: ``\texttt{yyyy-mm-dd: tweet or indicator value}''. 
Each part of the prompt begins and ends with \texttt{< >} and \texttt{</\ >} tags to enable easy post-processing. 
We instruct the model to generate a summarized analysis of the given input. We specify that the output will be used for predicting future indicator values. We explain about the input data and its structure. Finally, we break the task into two steps and instruct first to analyze the attributes that appear in the tweets and then analyze their potential effects on the close-future indicator values. 
A full example of a prompt is presented in Figure~\ref{fig:prompt}.

\textbf{LLM Parameters:} Initial experiments suggest that a temperature of 0.5 balances well between generating diverse and creative outputs while maintaining a factual analysis. All other parameters are the model's default.

\textbf{LLM prompt example}
\begin{figure}[H]
    \centering
    \begin{tcolorbox}[width=\textwidth]
        \begin{lstlisting}
You are a financial and NLP expert, assisting on creating a summarised analysis on textual and financial data.
Your task is to create an analysis on given tweets from Twitter and on S&P 500 values from the same time period.
Your output will be used for producing S&P 500 predictions in the close-future.
Your input 'Financial values' is a dictionary of S&P 500 values with their corresponding date (yyyy-mm-dd), from a time period of a month.
Your input 'Tweets' is a dictionary of tweets from Twitter with their publication date (yyyy-mm-dd), from the same time period. The tweets were posted by opinion leaders and discuss about the news, current affairs, economy, finance, and politics.
To produce this analysis, first analyse the fear, stability and stress expressed in the tweets, and then analyse their possible effects on the close-future S&P 500 value.
Produce your output in the format:  <Analysis of Tweets>PLACE_HOLDER</Analysis of Tweets>
<Potential Effects on S&P 500>PLACE_HOLDER</Potential Effects on S&P 500>

Input:
<Financial values>
2021-09-01: 4524.09
2021-09-02: 4536.95
...
2021-10-01: 4357.04
</Financial values>

<Tweets>
2021-09-01: It is time for a total economic boycott of Texas and Texas-based businesses.
...
2021-09-29: Pfizer is the 6th most owned stock by Congress. Surprised?
</Tweets>

Output:
        \end{lstlisting}
    \end{tcolorbox}
    \caption{LLM prompt for generating the summarized analysis of tweets and financial data of a monthly window.}
    \label{fig:prompt}
\end{figure}

\section{T5 Integration Prompt}\label{app:t5_prompt}
Figure \ref{fig:t5_prompt_example} shows a prompt example for the T5 model classifying ``direction change'' of the S\&P 500 for 1-day prediction horizon. This model is used for the experiments in Section \ref{subsec:llm_t5_exp}. The prompt contains both LLM-generated analysis and historical financial values, separated using leading titles and special tokens. The prompt ends with a title of the task itself. More details about the \textit{T} and \textit{F} integration can be seen in Section \ref{subsec:t_f_integration}.

\begin{figure}[H]
    \centering
    \begin{tcolorbox}[width=\textwidth]
        \begin{lstlisting}
[CLS] Summary of recent tweets: The tweets from the given time period cover a wide range of topics including politics, economy, finance, and social issues. There are tweets expressing concerns about the economy, such as discussions on inflation, high energy prices, and the impact of government policies on businesses. There are also tweets discussing the need for financial reforms, including calls for canceling student loan debt and increasing taxes on the wealthy and corporations. Additionally, there are tweets highlighting the consequences of defaulting on debt and the potential impact on the economy. Overall, the tweets reflect a mix of opinions and concerns regarding the current economic and financial landscape. [SEP] Recent S&P 500 directions of change: decrease=0, increase=1, increase=1, increase=1, decrease=0, decrease=0, decrease=0. [EOS] Predict S&P 500 direction of change tomorrow:</s>
    \end{lstlisting}
        \end{tcolorbox}
        \caption{Prompt example of the T5 prediction model receiving both LLM-generated analysis and historical financial values, in order to classify ``direction change'' of the S\&P 500.}
        \label{fig:t5_prompt_example}
\end{figure}

\section{Additional Narratives Analysis from Text}\label{app:textual_analysis}
This section presents complementary analyses to the ones presented in Section~\ref{sec:textual_analysis}.

\subsection{Twitter Analysis}\label{subapp:twitter_analysis}
\begin{figure}[h]
\centering
\includegraphics[width=0.6\columnwidth]{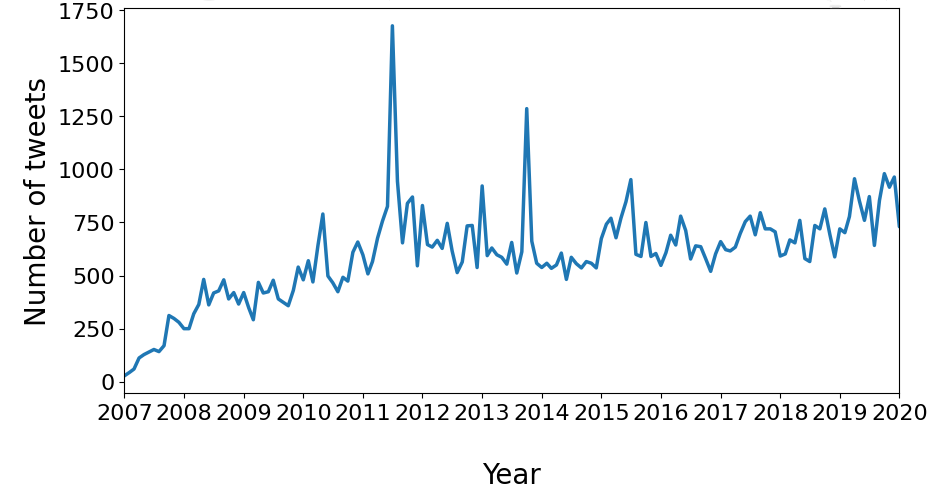}
\caption{Frequency of the word ``debt'' in the ``economics'' dataset over the years. Observed peaks in July 2011 and October 2013 are aligned with the major debt ceiling crises the US experienced in these times.}	
\label{fig:debt_occurences}	
\end{figure}

To inspect the information contained in our Pre-Pandemic Twitter dataset, we conduct an analysis of the 100 most frequent words associated with each topic of the dataset. This analysis reveals temporal shifts in the distribution of most frequent words, coinciding with significant world events. For instance, Figure \ref{fig:debt_occurences} illustrates the monthly number of occurrences of the term ``debt'' in the ``economics'' dataset, presenting peaks in July 2011 and October 2013, corresponding to major debt ceiling crises experienced by the US during these times.

\begin{figure*}[h]
\centering
\subfigure[Sentiment scores divided to negative and positive bins.]{
    \includegraphics[width=0.3\textwidth]{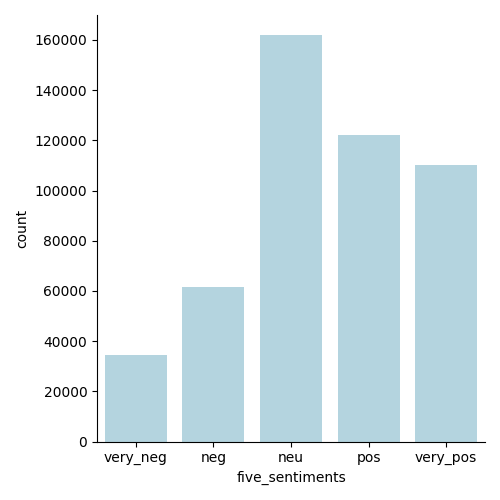}
    }
\vspace{10pt}
\subfigure[Raw sentiment scores.]{
    \includegraphics[width=0.48\textwidth]{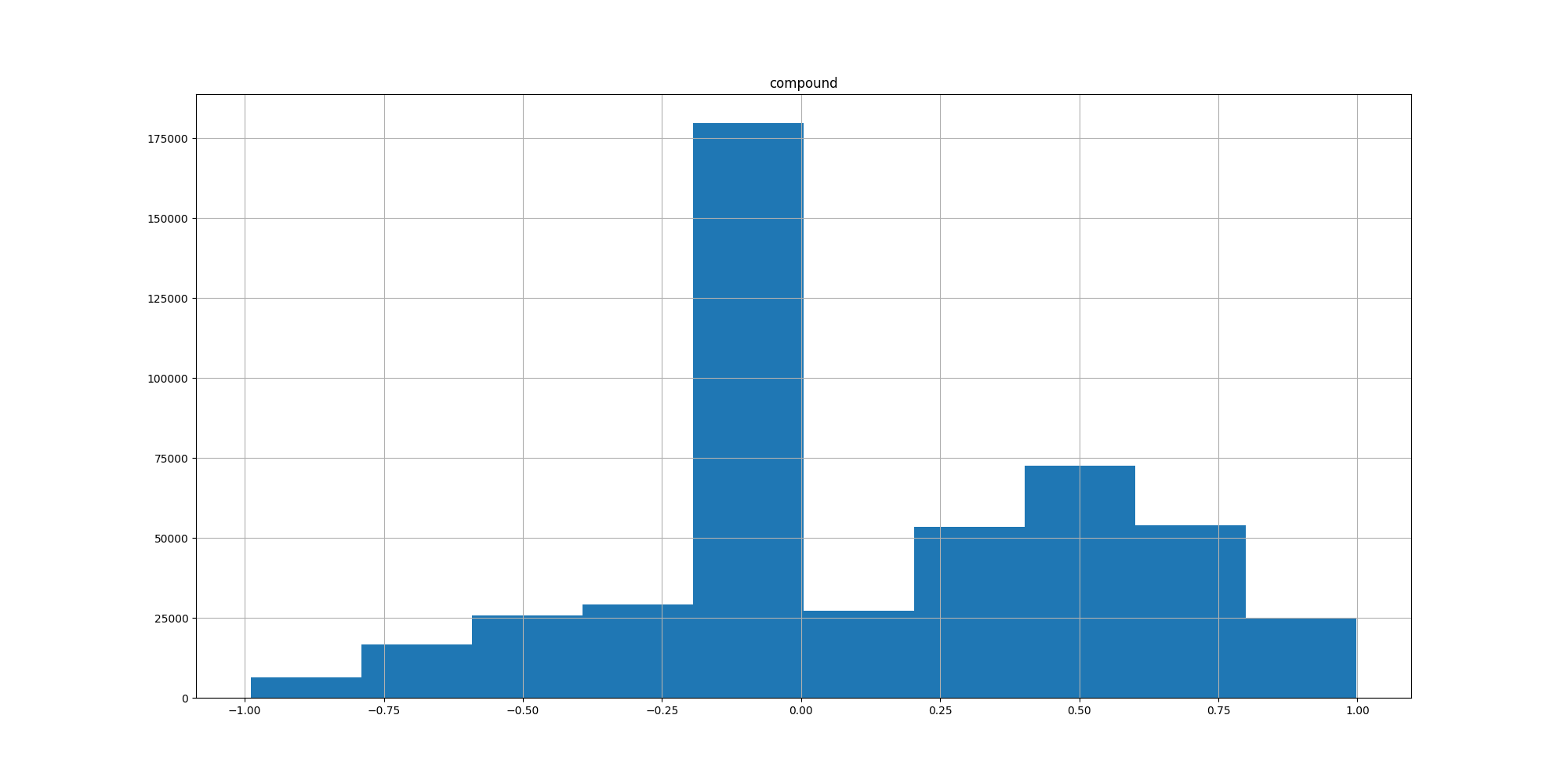}
    }
\caption{VADER \citep{hutto2014vader} sentiment scores of the ``business'' dataset.}	
\label{fig:vader_sentiment_scores}	
\end{figure*}

Figure~\ref{fig:vader_sentiment_scores} shows (a) sentiment scores created by VADER \citep{hutto2014vader} sentiment model (ranged [-1,1]) for the ``business'' related tweets; and (b) the sentiment scores divided to bins of size 0.5.


\subsection{LLM-Based Narrative Analyses Examples}\label{subapp:llm_analysis}
Figures \ref{fig:llm_analysis_example_1} and \ref{fig:llm_analysis_example_2} present LLM's outputs examples. Each analysis is constructed of several sentences describing the tweets and financial values given in the prompt. 
In both examples we can see the analyses recognize the described world events and analyze the impact of them on the financial activity, such as raising concerns and frustration, stress about the state of the economy market, or support in the government actions, which can impact the financial indicator.

\begin{figure}[!ht]
    \centering
    \begin{tcolorbox}[width=\textwidth]
        \begin{lstlisting}
Analysis of Tweets:
The tweets from the given time period cover a wide range of topics including Brexit, inflation, debt, unemployment, tax cuts, climate change, and business practices. There is a mix of opinions expressed, with some tweets criticizing government policies and economic decisions, while others highlight the need for investment in working families and fair economic opportunities.
Many tweets express concerns about the impact of Brexit on the UK, citing issues such as gas price hikes, food shortages, export crisis, and chaos in various industries. These tweets indicate a sense of fear and stress regarding the economic consequences of Brexit.
There are also tweets discussing the US economy, particularly the national debt and the potential consequences of defaulting on it. Some tweets criticize Republicans for their stance on the debt ceiling and the potential negative effects on jobs and unemployment. Other tweets highlight the need for tax reform and wealth redistribution to address income inequality.
Climate change and environmental issues are mentioned in several tweets, with some praising China's decision to stop building coal plants overseas and others criticizing the lack of action on climate change by certain politicians and corporations.
In terms of business practices, there are tweets discussing stock buybacks, corporate bailouts, and the influence of pharmaceutical companies in Congress. Some tweets call for increased regulation and transparency in the financial sector.
Overall, the tweets reflect a mix of concerns about the state of the economy, the role of government, and the need for equitable economic policies.

Potential Effect on S&P 500:
Based on the analysis of the tweets, it is evident that there are several factors that could potentially impact the S&P 500 in the close future. These factors include:
1. Brexit-related concerns: The tweets expressing fear and stress about the economic consequences of Brexit could lead to increased market volatility and uncertainty, which may negatively impact the S&P 500.
2. Debt ceiling and default risks: The tweets discussing the potential consequences of defaulting on the national debt highlight the risks to the economy and job market. If the debt ceiling issue is not resolved in a timely manner, it could lead to market instability and potentially impact the S&P 500.
3. Climate change and environmental policies: The tweets mentioning climate change and the need for clean energy transition may have implications for certain sectors of the economy, such as fossil fuels and renewable energy. Changes in government policies and regulations related to climate change could impact the performance of companies in these sectors, which in turn could affect the S&P 500.
4. Business practices and regulations: The tweets criticizing stock buybacks, corporate bailouts, and the influence of pharmaceutical companies may contribute to calls for increased regulation and scrutiny of these practices. Any regulatory changes or increased scrutiny on these issues could impact the performance of companies involved, potentially affecting the S&P 500.
It is important to note that the impact of these factors on the S&P 500 will depend on various other economic and market conditions. However, considering the sentiments expressed in the tweets, it is advisable to closely monitor these factors and their potential effects on the S&P 500 in the close future.
        \end{lstlisting}
    \end{tcolorbox}
    \caption{LLM-generated analysis for dates 21/09/2021 to 21/10/2021. In this time period the US Labor Department reported that the unemployment rate fell to the lowest level since the start of the COVID-19 pandemic, the US Federal Reserve announced that it would begin tapering its asset purchases, and the US House of Representatives passed the Build Back Better Act (a bill for investments in climate change, healthcare, education and child care) and a week after the US senate failed to pass it.}
    \label{fig:llm_analysis_example_1}
\end{figure}

\begin{figure}[H]
    \centering
    \begin{tcolorbox}[width=\textwidth]
        \begin{lstlisting}
Analysis of Tweets:
The tweets cover a wide range of topics including politics, finance, economy, and current affairs.
There are tweets discussing the actions and statements of political figures such as Marco Rubio, Donald Trump, and Boris Johnson.
Some tweets express concerns about inflation, rising interest rates, and the impact on the economy and personal finances.
There are mentions of financial fraud, insider trading, and the role of big corporations in driving inflation.
Several tweets highlight the impact of government policies on various sectors, such as healthcare, student loans, and housing.
Some tweets express skepticism towards central banks and their role in the economy.
There are mentions of potential economic crises, recessions, and the need for fiscal stimulus.
A few tweets discuss the impact of global events, such as the Russian mobilization and the European Central Bank's digital euro.
Some tweets express frustration with the current state of the economy and the financial hardships faced by individuals.

Potential Effect on S&P 500:
- The tweets reflect a mix of positive and negative sentiments towards the economy and financial policies. This sentiment can influence investor confidence and market sentiment, potentially impacting the S&P 500.
- Concerns about inflation, rising interest rates, and economic instability expressed in the tweets may lead to increased market volatility and potential declines in the S&P 500.
- Tweets discussing financial fraud, insider trading, and negative perceptions of big corporations may erode investor trust and confidence, potentially impacting the S&P 500.
- The mention of potential economic crises, recessions, and the need for fiscal stimulus may create uncertainty and caution among investors, potentially affecting the S&P 500.
- Global events mentioned in the tweets, such as the Russian mobilization and the European Central Bank's digital euro, may have indirect effects on the S&P 500 through their impact on global markets and investor sentiment.
- Overall, the sentiment and concerns expressed in the tweets suggest a cautious outlook towards the economy and financial markets, which may influence S&P 500 predictions in the close future.
        \end{lstlisting}
    \end{tcolorbox}
    \caption{LLM-generated analysis for dates 29/08/2022 to 28/09/2022. In this time period the Federal Reserve raised the interest rates in an effort to combat inflation, the US Supreme Court ruled that the Biden administration could not extend the pause on student loan payments, the US unemployment rate fell to the lowest level since 1969, and the Biden administration announced an aid package for Ukraine.}
    \label{fig:llm_analysis_example_2}
\end{figure}

\section{Full Results for ``Sentiment-Based Next-Day Prediction'' Experiment}\label{app:sent_exp}

This appendix shows the full results for the experiment described in Section \ref{subsec:sentiment_exp}. 
Tables \ref{tab:full_vader_ml_cls} and \ref{tab:full_vader_ml_reg} presents the performance of all compared models for both ``direction change'' and ``next value'' tasks. 
For the summary of results and takeaways, please refer to the main experiment section.

\begin{table}[H]
\small  
\setlength{\tabcolsep}{8pt}  
\begin{tabularx}{\columnwidth}{l|l|c} 
\toprule
\textbf{Type} & \textbf{Model}& \textbf{Accuracy}\\
\midrule
F- & As-previous & 0.812\\
baselines & Train-majority & 0.885\\
\hline
F & Logistic Regression Binary & 0.884 \\
F & Logistic Regression Numeric & 0.896 \\
F & SVM Binary & 0.883\\
F & SVM Numeric & 0.899\\
F & Random Forest Binary & 0.878\\
F & Random Forest Numeric & \bfseries{0.936}\\
\hline
TF & Logistic Regression Binary & 0.884 \\
TF & Logistic Regression Numeric & 0.895 \\
TF & SVM Binary & 0.883\\
TF & SVM Numeric & 0.899 \\
TF & Random Forest Binary & 0.880\\
TF & Random Forest Numeric & \bfseries{0.939}\\
\hline
T & Logistic Regression & 0.885\\
T & SVM & 0.885\\
T & Random Forest & 0.879\\
\bottomrule
\end{tabularx}
\caption{``Direction change'' classification of FFR. Financial features are encoded either as ``Binary'' (increase/decrease) or ``Numeric''.
The TF and \textit{F} models perform nearly identically, outperforming the \textit{T} models and the \textit{F} baselines. Best performing models are \textit{F} \& \textit{TF} Random Forest with numeric financial features.}
\label{tab:full_vader_ml_cls}
\end{table}

\begin{table}[h!]
\small  
\setlength{\tabcolsep}{8pt}  
\begin{tabularx}{\columnwidth}{l|l|c}
\toprule
\textbf{Type} & \textbf{Model} & \textbf{MSE} \\ 
\midrule
F- & As-previous & 28.712 \\ 
baselines & Train-mean & \bfseries{15.661} \\ 
\hline
F & Linear Regression &  18.842 \\ 
F & Lasso &  18.135\\ 
F & Ridge & 18.842 \\ 
F & SVM & \bfseries{15.416} \\ 
F & Random Forest & 37.081 \\ 
\hline
TF & Linear Regression & 19.152 \\ 
TF & Lasso & 18.135 \\ 
TF & Ridge & 19.121 \\ 
TF & SVM & \bfseries{15.416} \\ 
TF & Random Forest & 37.576 \\ 
\hline
T & Linear Regression & 15.932 \\ 
T & Lasso & 15.661 \\ 
T & Ridge & 15.898 \\ 
T & SVM & \bfseries{15.36} \\ 
T & Random Forest & 24.629 \\ 
\bottomrule
\end{tabularx}
\caption{``Next value'' prediction of FFR. Best \textit{T}, \textit{F}, \textit{TF} models are comparable to each other and to ``train-mean'' \textit{F} baseline.} 
\label{tab:full_vader_ml_reg}
\end{table}


\section{Full Description for ``LLMs for End-to-End Prediction'' Experiment}\label{app:llm_as_predictor}
This appendix is complementary to Section \ref{subsec:llm_exp}.

\textbf{Method.} Motivated by the limitations observed in prior experiments, we explore the utility of LLMs in predicting the VIX index by utilizing their ability to process long textual sequences and perform multi-step reasoning, obviating the need for separate components for text analysis and financial prediction. We utilize the web chat version of GPT \citep{gpt-chat} for continuous reasoning.

We predict the average VIX value for the following week based on a monthly window of tweets paired with corresponding VIX values. Formally, given data covering dates $T_1,...,T_{30}$, the LLM is instructed to predict the average value of the week between $T_{30+7} - T_{30+14}$ (horizon=7). This target mitigates random daily fluctuations common in economic indicators, enabling the model to capture broader trends.

Notably, similar behaviors were seen for S\&P 500 prediction, and are omitted due to space constraints.

\textbf{1. Few-Shot Learning:} Initially, we condition the LLM with examples of tweet-VIX pairs, instructing it to generate similar predictions for unseen inputs.

\textbf{Results.} While the model provides descriptive analyses of inputs, although not being instructed to do so, it faces challenges in financial prediction. It sometimes refuses to provide predictions, citing its limitations as a language model and its inability to provide financial advice or prediction. In other cases it predicts a range of values mirroring the input range, neglecting potential effects of current tweets' narratives which it successfully analyzes.

\textbf{2. Multi-Step CoT Reasoning:} To address the above challenges, we adopt a multi-step approach using chain-of-thought (CoT) instructions. The LLM is instructed to first analyzing tweets, then explaining their potential influence on the VIX, and finally predicting the average VIX for the following week, while providing a rational for the prediction. This unified approach aims to encourage the model to generate analyses directly relevant to the final prediction task.


\textbf{Results.} The LLM produces multi-step outputs, demonstrating meaningful narrative extraction and comprehension of their financial implications. However, prediction performance is inconsistent, with occasionally producing reasonable predictions and explanations, alongside instances where previous mentioned challenges still arise.

\begin{figure*}[!htb]
\centering
\captionsetup{skip=1pt} 
\subfigure[Trend Reversal Period.]{
    \includegraphics[width=0.48\textwidth]{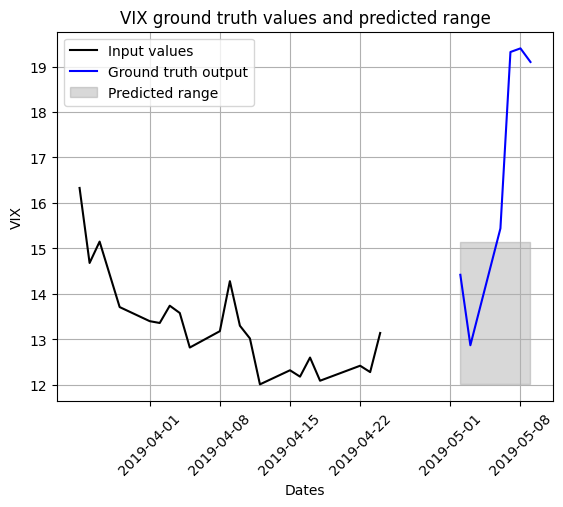}
    }
\subfigure[Volatile period.]{
    \includegraphics[width=0.48\textwidth]{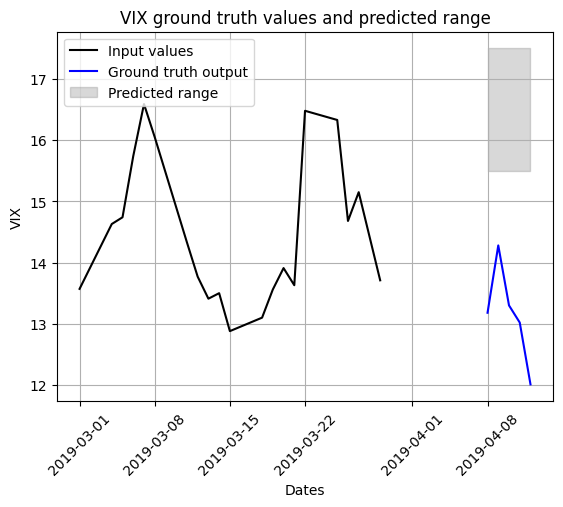}
    }
\caption{VIX economic definite periods, together with the LLM's predicted ranges.}	
\label{fig:vix_economic_periods}	
\end{figure*}

\textit{Analyzing Economic Periods:} Since the financial indicator can be volatile during a monthly period, we assess the LLM's performance during stable (declining VIX), stressed (increasing VIX), trend reversal (declining present and increasing future-surprising peak), and volatile (fluctuating VIX) periods, assessing its prediction abilities in definite periods. We compare predictions with and without tweet data (\textit{F} vs. \textit{TF} models). 
Figure~\ref{fig:vix_economic_periods} presents the VIX values during such periods, together with the LLM's predicted range.


\textit{Stable Period:} Both \textit{F} and \textit{TF} models correctly predict a continued decline, with the \textit{TF} model providing richer analysis.


\textit{Stressed Period:} The LLM fails to capture the upward trend, despite identifying stress in tweets and a current moderate-high VIX values. Revises predictions upon prompting favor the higher end of the current value range.


\textit{Trend Reversal:} Despite recognizing the declining trend and increased stress, the LLM predicts a continuation of the declining trend, indicating limitations in incorporating dynamic changes.


\textit{Volatile Period:} Both \textit{F} and \textit{TF} models predict wide ranges, mirroring the input, with the \textit{TF} model offering richer analysis, capturing emotional cues, economic discussions, and political events, all potentially contributing to volatility. While it suggests a slight increase, the wide range limits its precision.

\textit{Multi-Step Few-Shot:} This periods analyses reveal that the textual data seems to enrich the LLM's understanding, leading to more nuanced explanations but only occasionally offering tailored numerical predictions. To address this, we revisited the few-shot setting, providing the LLM with both illustrative input-output pairs and multi-step instructions. This aims to quantify the magnitude and direction of tweets' narratives influence on future financial values. Yet, the LLM's predicted ranges remain incongruent with its tweets' analysis.


\textit{Synthetic Scenarios:} We manipulate input data to investigate the LLM's grasp of causal relationships, feeding it with texts coupled with corresponding financial changes (see the ``synthetic narratives'' baseline in \S~\ref{subsec:textual_baselines}).
While the LLM recognizes the direction of impact of these scenarios, it struggles to quantify the actual magnitude of the change.


\textbf{Takeaway.} The LLM shows potential for financial forecasting, with its ability to analyze textual data and identify potential economic impacts, offering promising avenues for market insights. Yet, the model faces challenges in prediction reluctance and inconsistency in incorporating textual insights. Addressing these issues can unlock the full potential of LLMs for robust and narratives-insightful financial forecasting.


\end{document}